\documentclass[10pt,twocolumn,letterpaper]{article}

\usepackage[pagenumbers]{cvpr} 
\usepackage{xcolor}
\usepackage{amsmath}
\usepackage{amssymb}
\usepackage{array}
\usepackage{multirow}
\usepackage{color}
\usepackage{colortbl}
\usepackage{framed}
\usepackage{bm}
\usepackage{bbm}
\usepackage{xspace}
\usepackage{enumitem}
\usepackage{algorithm}
\usepackage{listings}
\usepackage{url}
\usepackage{booktabs}
\usepackage{pifont} 

\definecolor{Gray}{gray}{0.9}
\definecolor{LightCyan}{rgb}{0.88,0.95,1}
\definecolor{blond}{rgb}{0.98, 0.94, 0.75}
\definecolor{green}{rgb}{0.0, 0.62, 0.42}
\definecolor{lightgray}{rgb}{0.83, 0.83, 0.83}

\newcommand{\cmark}{\ding{51}}%
\newcommand{\xmark}{\ding{55}}%

\newcommand{\ours}{FreeDA\xspace}

\newcommand{\tit}[1]{\smallbreak\noindent\textbf{#1.}}
\newcommand{\tinytit}[1]{\noindent\textbf{#1.}}

\newcommand\blfootnote[1]{%
  \begingroup
  \renewcommand\thefootnote{}\footnote{#1}%
  \addtocounter{footnote}{-1}%
  \endgroup
}

\definecolor{cvprblue}{rgb}{0.21,0.49,0.74}
\usepackage[pagebackref,breaklinks,colorlinks,citecolor=cvprblue]{hyperref}

\def\paperID{7500} 
\def\confName{CVPR}
\def\confYear{2024}

\title{Training-Free Open-Vocabulary Segmentation\\with Offline Diffusion-Augmented Prototype Generation}
\author{Luca Barsellotti$^{1\hspace{0.02cm}*}$\quad Roberto Amoroso$^{1\hspace{0.02cm}*}$ \quad Marcella Cornia$^1$ \quad Lorenzo Baraldi$^1$ \quad Rita Cucchiara$^{1,2}$\\
$^1$University of Modena and Reggio Emilia, Italy \quad $^2$IIT-CNR, Italy\\
{\tt\small \{name.surname\}@unimore.it}
}

\begin{document}
\maketitle

\begin{abstract}

Open-vocabulary semantic segmentation aims at segmenting arbitrary categories expressed in textual form. Previous works have trained over large amounts of image-caption pairs to enforce pixel-level multimodal alignments. However, captions provide global information about the semantics of a given image but lack direct localization of individual concepts. Further, training on large-scale datasets inevitably brings significant computational costs.
In this paper, we propose \ours, a training-free diffusion-augmented method for open-vocabulary semantic segmentation, which leverages the ability of diffusion models to visually localize generated concepts and local-global similarities to match class-agnostic regions with semantic classes. Our approach involves an offline stage in which textual-visual reference embeddings are collected, starting from a large set of captions and leveraging visual and semantic contexts. At test time, these are queried to support the visual matching process, which is carried out by jointly considering class-agnostic regions and global semantic similarities.
Extensive analyses demonstrate that \ours achieves state-of-the-art performance on five datasets, surpassing previous methods by more than 7.0 average points in terms of mIoU and without requiring any training. Our source code is available at \href{https://aimagelab.github.io/freeda/}{aimagelab.github.io/freeda}. 
\blfootnote{$^*$Equal contribution.}
\end{abstract}    
\section{Introduction}
\label{sec:intro}

Semantic segmentation is a core problem in Computer Vision, which aims at partitioning an image into coherent regions according to a set of semantic categories~\cite{long2015fully,noh2015learning}. As manually annotating large-scale amounts of training data is expensive, scaling segmentation to large sets of concepts in a fully supervised manner is impracticable. This has recently moved the focus of the community towards open-vocabulary solutions~\cite{zhao2017open,ghiasi2022scaling,ding2022decoupling,liu2022open,li2022languagedriven,xu2022simple} that, learning from a narrow set of seen categories or weak forms of supervision, are able to segment novel and unseen categories.

\begin{figure}[t]
    \centering
    \includegraphics[width=0.99\linewidth]{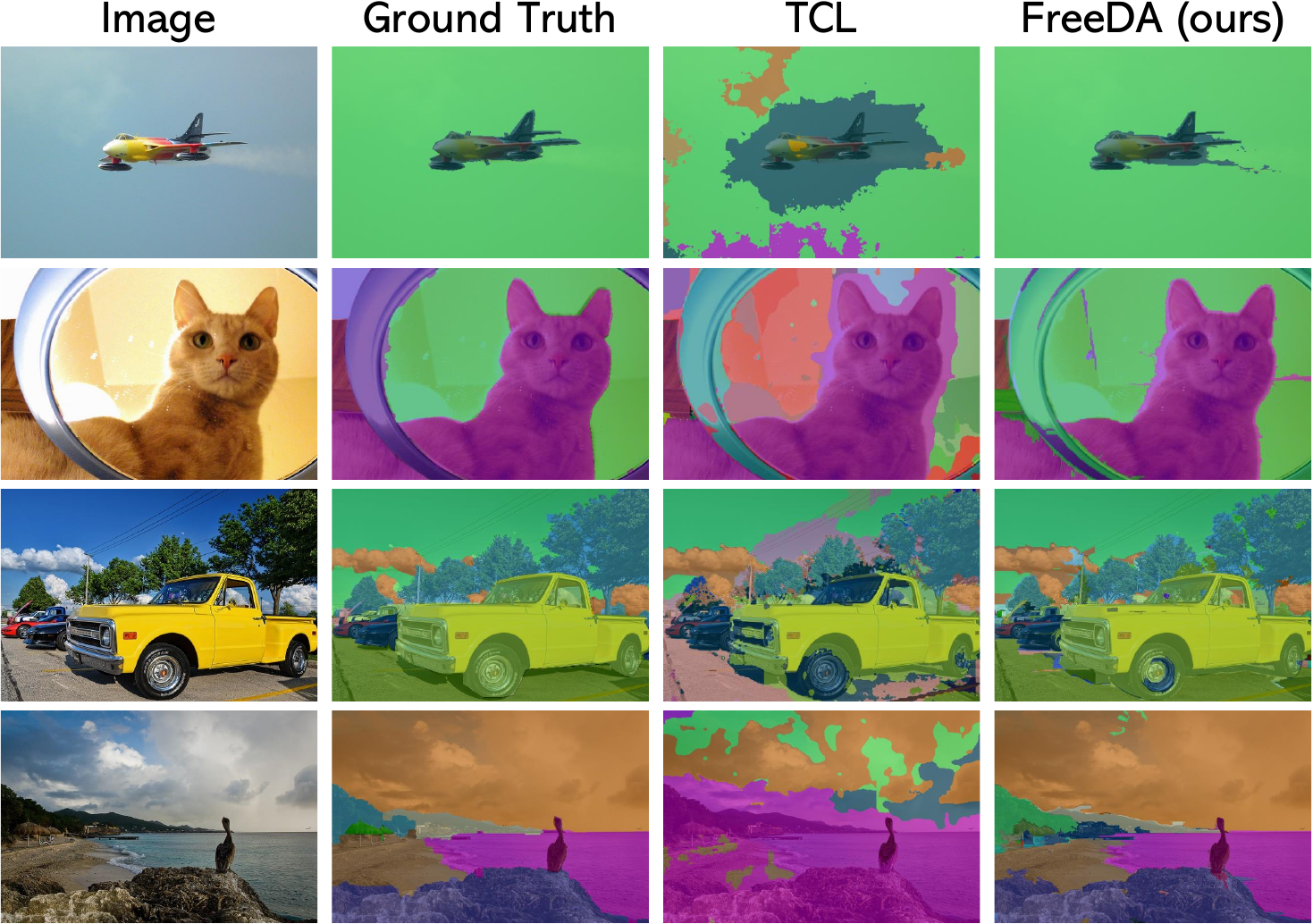}
    \vspace{-0.1cm}
    \caption{Open-vocabulary segmentation with: (a) TCL~\cite{cha2023learning}, which performs end-to-end learning of region-text alignment; (b) our \ours, which leverages generated textual-visual embeddings with global-local similarities and does not require any training.}
    \label{fig:first_page}
    \vspace{-0.35cm}
\end{figure}

One of the major challenges in this setting is how to transfer the ability to match texts and images of large-scale vision-language models (\eg, CLIP~\cite{radford2021learning} and ALIGN~\cite{jia2021scaling}) to a text-pixel alignment. Given a large-scale set of web-crawled image-caption pairs, previous approaches~\cite{cha2023learning,xu2022groupvit,xu2023learning,zhou2022extract,ren2023viewco,luo2023segclip} force the ability to localize textual concepts to emerge through contrastive learning techniques combined with grounding mechanisms~\cite{xu2022groupvit,xu2023learning,cha2023learning}. However, captions often capture the global scene and might present ambiguities with respect to fine-grained elements, making this approach sub-optimal and computationally intensive.

On a different note, advances in diffusion models~\cite{rombach2022high,ho2022cascaded} have shown remarkable results in text-to-image generation, and recent works have shown that their features encompass knowledge regarding the positioning of the generated objects~\cite{tang2023daam,wu2023diffumask,li2023grounded}.
This information can be exploited to generate large sets of attribution maps, which are more active in the area corresponding to a semantic class, thus providing a valuable source of information for semantic segmentation. We propose to explore this mechanism as an alternative to multimodal contrastive training, in a fully training-free methodology where no parameter is learned.

In contrast to previous works, our proposed approach follows an efficient two-step protocol: in an offline stage, we leverage a diffusion-augmented generation in which a collection of textual-visual reference vectors is generated. Then, at inference time, these references are retrieved to compute local and global similarities to segment the input image. In detail, we employ a large set of textual captions to generate synthetic images and corresponding attribution maps, through a localization mechanism based on cross-attention. Subsequently, we leverage a self-supervised visual backbone, DINOv2~\cite{oquab2023dinov2}, to build an offline set of visual prototypes associated with textual vectors, each representing the context of an instance in its synthetic scene.

At inference time, we extract both global features with a multimodal encoder (\ie, CLIP) and local dense features with DINOv2, characterized by high semantic relatedness, and employ a superpixel algorithm to detect class-agnostic regions. By querying the input textual category in the set of textual-visual reference embeddings, we then assign each superpixel to the category that exhibits the highest combined similarity, between the global and local modalities. As our approach is training-\textit{free} and relies on \textit{D}iffusion-\textit{A}ugmented generation, we name it \ours.

We validate the proposed framework by conducting extensive experiments on Pascal VOC~\cite{pascal-voc-2012}, Pascal Context~\cite{mottaghi2014role}, COCO Stuff~\cite{caesar2018coco} and Object~\cite{lin2014microsoft}, Cityscapes~\cite{cordts2016cityscapes}, and ADE20K~\cite{zhou2017scene,zhou2019semantic}. Without requiring any form of training, \ours consistently outperforms previous approaches by a large margin, achieving state-of-the-art performance on all datasets. Overall, our work demonstrates that non-parametric approaches can provide a compelling and efficient alternative for open-vocabulary semantic segmentation, and opens up new opportunities for subsequent works.
To sum up, the contributions of this paper are as follows:
\begin{itemize}
    \item We introduce FreeDA, a novel training-free method for open-vocabulary semantic segmentation based on the generation of context-aware textual-visual reference embeddings through diffusion models.
    \item We present an inference pipeline that, leveraging the semantic correspondence of DINOv2, superpixel algorithms, and a combination of local and global similarities achieves precise and robust segmentation prediction.
    \item Our experiments show that our approach achieves state-of-the-art performance on five datasets, without requiring any form of training.
\end{itemize}
\section{Related Work}
\label{sec:related}

\tit{Open-Vocabulary Semantic Segmentation} Building upon the success of large-scale vision-language models in zero-shot classification~\cite{jia2021scaling,radford2021learning}, previous works on open-vocabulary segmentation have investigated strategies to transfer the multimodal image-text alignment toward finer granularity (\ie, region or pixel level)~\cite{ghiasi2022scaling,ding2022decoupling,liang2023open,xu2023side,xu2022groupvit}.

A group of literature has been focusing on the supervision provided by dense annotations, available for a limited set of categories, to generalize on unseen classes. OpenSeg~\cite{ghiasi2022scaling} decouples the task in a region proposer and a grounder that aligns regions to words from captions. Similarly, OVSeg~\cite{liang2023open} employs a two-stage method, in which class-agnostic regions are masked and provided to a CLIP encoder with learnable visual prompts. SAN~\cite{xu2023side} combines a side network with CLIP to propose regions while recognizing their corresponding semantic category.
However, these approaches are affected by performance gaps between seen and unseen categories~\cite{ding2022decoupling,liang2023open} and, due to the costs of dense annotations, can be applied in limited domains.

Other works have instead exploited contrastive training over a large set of image-text pairs, without dense annotations. GroupViT~\cite{xu2022groupvit} proposes a Transformer architecture that learns to group image regions progressively. MaskCLIP~\cite{zhou2022extract} adapts a frozen CLIP for dense predictions through modifications in the last attention layer. TCL~\cite{cha2023learning} presents a grounding mechanism that learns to associate text to regions during contrastive learning. OVSegmentor~\cite{xu2023learning} introduces a module based on slot attention to group tokens of a Transformer and aligns them to captions.
Our approach falls into this research direction, since it relies only on a set of captions as support, without requiring dense annotations.

\tit{Localization in Diffusion Models} Diffusion models~\cite{rombach2022high} have proven state-of-the-art performance in image generation.
Few works tackle the task of localizing the concepts mentioned in the conditioning captions during the generation. DAAM~\cite{tang2023daam} proposes exploiting the cross-attention mechanism that Stable Diffusion uses to extract attribution maps for the words mentioned in the prompt. DiffuMask~\cite{wu2023diffumask} leverages the advances of DAAM to generate ground truth segmentation masks without human annotation and train a segmentation model on them. GroundedDiffusion~\cite{li2023grounded} implements a grounding module to align textual and visual embeddings during the diffusion process. 

Some works have investigated the usage of diffusion models for open-vocabulary segmentation. ODISE~\cite{xu2023open} employs Stable Diffusion as a feature extractor for its mask generator. OVDiff~\cite{karazija2023diffusion} generates a set of visual references at prediction time to support the segmentation process. Our approach also relies on the generation of images; however, this is done to collect visual prototypes during an offline stage, a choice that significantly reduces the computational load at prediction time.

\tit{Superpixel Algorithms}
The concept of superpixel arises from the observation that pixels are not a natural representation of an image. A superpixel is a group of homogeneous pixels based on the visual characteristics of the image, such as shape, brightness, color, and texture. Over the years, several extraction strategies have been developed with the goal of improving their quality and efficiency, such as watershed-based~\cite{hu2015watershed,machairas2014waterpixels,neubert2014compact} and clustering-based~\cite{achanta2010slic,achanta2012slic,li2015superpixel} approaches. In this paper, we employ superpixels as a support for partitioning the image into class-agnostic regions, from which local visual similarities are computed.
\section{Proposed Method}
\label{sec:method}

\begin{figure}[t]
    \centering
    \includegraphics[width=0.85\linewidth]{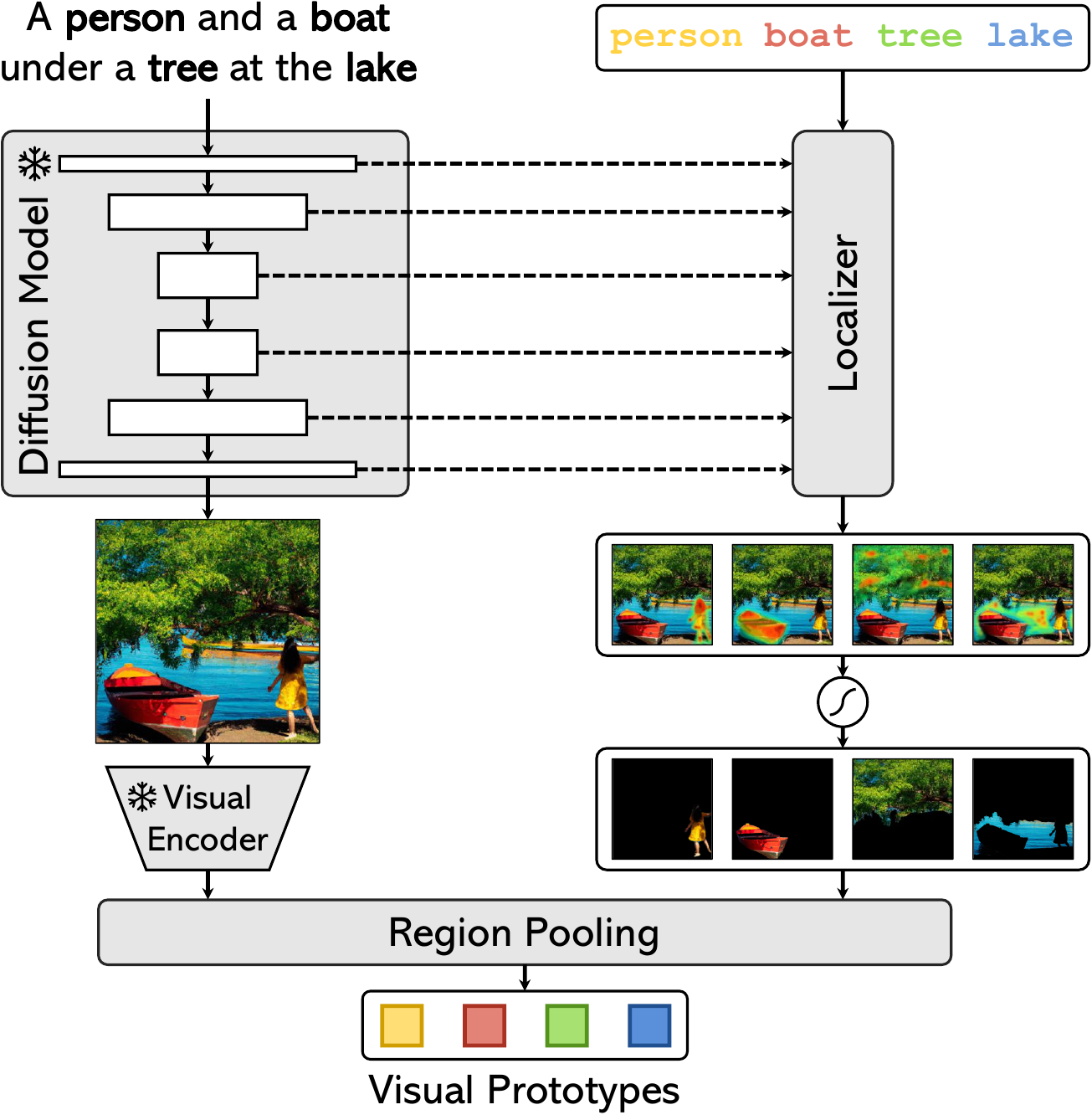}
    \vspace{-0.15cm}
    \caption{Overview of the diffusion-augmented prototype generation phase of \ours. Visual prototypes are generated by pooling self-supervised visual features on weak localization masks extracted from Stable Diffusion.}
    \label{fig:prototype_extraction}
    \vspace{-0.35cm}
\end{figure}

\begin{figure*}[t]
    \centering
    \includegraphics[width=0.98\linewidth]{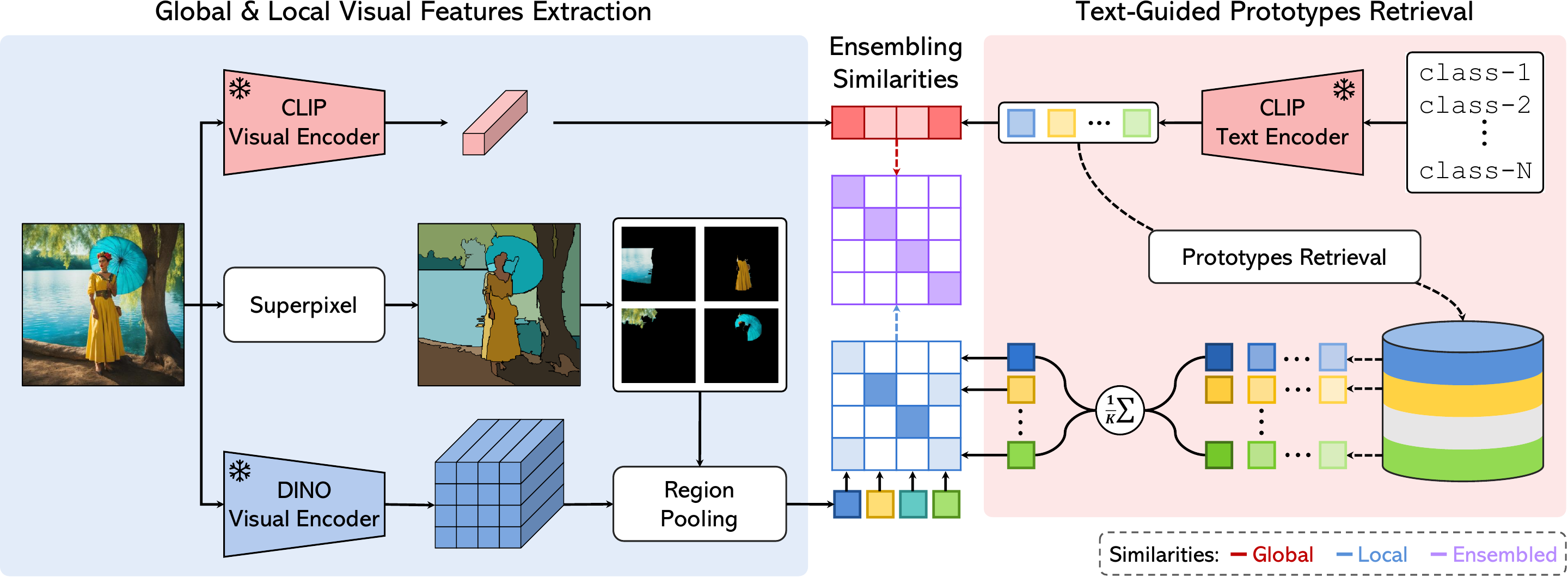}
    \vspace{-0.1cm}
    \caption{Overview of the inference process in \ours. Local (region-level) and global similarities are computed by employing, respectively, visual self-supervised and multimodal contrastive embedding spaces, and by comparing them with input texts and prototypes, built during the off-line stage.}
    \label{fig:inference}
    \vspace{-0.3cm}
\end{figure*}

The goal of open-vocabulary segmentation is to segment an image according to an arbitrary set of categories represented through free-form texts. Our training-free approach decouples the task into two phases: a \textit{diffusion-augmented prototype generation} phase, which is carried out in an off-line manner (visually represented in Figure~\ref{fig:prototype_extraction}), and a \textit{semantic correspondence-based inference} stage, which is employed at test time to perform prediction over an input image. This second stage is visually depicted in Figure~\ref{fig:inference}.

\subsection{Diffusion-Augmented Prototype Generation}
\label{sec:prototype-generation}
During the pre-processing phase, we collect a large set of visual prototypes and corresponding textual key embedding vectors, which describe semantic instances along with their textual and visual contexts. A textual key represents a semantic category and its textual context as described in a caption. A visual prototype, instead, describes an instance of that semantic category contextualized in an image. Collections of prototypes belonging to the same semantic class, thus, represent examples of the visual variety of that class.

\tit{Extracting Localized Masks with Diffusion Models}
As prototypes will be employed to predict semantic classes in a non-parametric way, it is crucial to build a large collection of prototypes with high semantic variance. 
To this aim, we generate a large set of real-world scenes using Stable Diffusion~\cite{rombach2022high} starting from a large set of captions. Generating images rather than collecting real images from web-scale datasets allows us to control the resulting semantic distribution and its variance. Most importantly, also, latent-based diffusion models can predict the location of objects in the generated scene~\cite{tang2023daam}. 

Diffusion models, indeed, map word embeddings of the conditioning text to the activations of their denoising subnetwork (\eg, U-Net~\cite{rombach2022high,ronneberger2015u}) through cross-attention layers applied at different scales. Cross-attention activations, therefore, relate each word of the conditioning caption to a portion of the image and can be employed to generate weak localization masks. As each layer of the denoising network produces cross-attention maps at a different scale, we upscale all intermediate maps at the original image size. Then, we collapse across heads, layers, and diffusion time steps to obtain a single object mask. 

Formally, the attribution map of a word $w$ from the conditioning caption over a generated image $I$ is expressed as
\begin{equation}
    A(I, w) = \frac{1}{T L H}\sum_{t, l, h} \text{upsample}(\mathcal{A}(I, w)_{t,l,h}),
\end{equation}
where $\mathcal{A}(I, w)$ indicates the collection of cross-attention maps with respect to the tokens of word $w$, and $t$, $l$, and $h$ index diffusion time steps, denoising layers, cross-attention heads respectively. Finally, $\text{usample}(\cdot)$ denotes a bilinear interpolation operator.

With the aforementioned approach for building localized masks, we employ a set of captions, designed to describe real images, to condition Stable Diffusion~\cite{rombach2022high} and generate the corresponding set of synthetic images. Through a noun parser~\cite{loper2002nltk}, from each caption we also extract mentioned nouns $\{w_1, \dots, w_N\}$ and obtain their corresponding attribution maps $A(I,w_i) \in \mathbb{R}^{H \times W}$ over the generated image. Then, we normalize the scores of the attribution maps in the range $[-1, 1]$, apply a sigmoid function, and binarize the result by thresholding it to a constant value $\gamma$. The output of this process is a weak localization mask $M(I, w_i) \in \{0, 1\}^{H \times W}$ for each noun $w_i$ mentioned in the input caption.

\tit{Visual Prototypes Extraction}
To encode the content of the aforementioned weak localization masks, we adopt DINOv2~\cite{oquab2023dinov2}, which showcases good localization and semantic matching capabilities. Given a generated image $I \in \mathbb{R}^{H \times W \times 3}$, we extract its dense features $v(I) \in \mathbb{R}^{\frac{H}{P} \times \frac{W}{P} \times d_v}$, where $P$ is the input patch size of the backbone and $d_v$ is the dimensionality of its embedding space. For every noun $w_i$ in the sentence, we interpolate the weak localization mask $M(I, w_i)$ to the size of the dense features, obtaining a resized version of the localization mask $\hat{M}(I, w_i) \in \mathbb{R}^{\frac{H}{P} \times \frac{W}{P}}$. Then, we perform a region pooling operation to aggregate visual features over the localization mask, as follows:
\begin{equation}
\label{eq:region_pooling}
    p(I, w_i) = \frac{\sum_{h=0}^{\frac{H}{P}} \sum_{w=0}^{\frac{W}{P}} v(I)[h, w] \hat{M}(I, w_i)[h, w]}{\sum_{h=0}^{\frac{H}{P}} \sum_{w=0}^{\frac{W}{P}} \hat{M}(I, w_i)[h, w]},
\end{equation}
where square brackets indicate indexing over spatial axes. The resulting vector $p(I, w_i) \in \mathbb{R}^{D_v}$ is the \textit{visual prototype} for the noun $w_i$ extracted from the input image $I$, and is defined as the mean of the dense features covered by the corresponding binary mask. Prototypes built with this approach embed a visual descriptor of the corresponding word localized in a synthetic context, obtained from a real description.

\tit{Textual Keys Extraction}
In addition to representing visual prototypes, we employ a text encoder to represent nouns in their lexical context. To this aim, we define a set of textual templates $\mathcal{T}$ (\eg, \texttt{A photo of a [NOUN]}), and embed each noun in all templates. This results in a textual embedding for each template, $t_{i}(w) \in \mathbb{R}^{D_t}, i=1 \dots, T$, where $T$ is the number of templates. We define $\hat{t}(w) = \frac{\sum_{i=1}^{T} t_{i}(w)}{T}$ as the mean noun embedding, and then linearly interpolate with the full caption embedding $\hat{c}$ to also capture the global context of the entire scene. Specifically, the resulting textual key vector $k(c, w)$ for a word $w$ taken from a caption $c$ is then defined as
\begin{equation}
    k(c, w) = \alpha \hat{t}(w) + (1-\alpha) \hat{c},
\end{equation}
where $\alpha \in (0,1)$ is a scalar weight. Similar to prototypes, keys obtained through this process represent nouns contextualized in the caption in which they have been extracted. As each textual key is associated with a visual prototype, the set of textual keys extracted from a dataset can be indexed via an approximate nearest neighbor search to efficiently retrieve visual prototypes given a textual query.

\subsection{Training-Free Mask Prediction}
At inference time, our goal is to query the keys of the pre-built collection index to retrieve their corresponding prototypes. Then, we employ these prototypes as references to segment the input image through semantic correspondence with both local and global features.

\tit{Retrieving Prototypes}
Given a set of textual categories $\{c_1, \dots, c_S\}$, we consider the same set of templates $\mathcal{T}$ employed during textual keys computation and embed each category as $\hat{t}(c_i) = \frac{\sum_{j=1}^T t_{j}(c_i)}{T}$, where $t_{j}(c_i)$ is the text embedding of a template applied on a category. For each category $c_i$, we leverage $\hat{t}(c_i)$ to query the key embeddings of the pre-built collection index and retrieve the $K$ most similar ones according to cosine similarity. 
Each key embedding corresponds to the combination of the text embeddings of both a noun and the caption in which the noun is mentioned, and is uniquely linked with a visual prototype. 
Hence, we compute a representative visual prototype for each category as the mean of retrieved prototypes. Formally, 
\begin{equation}
    \overline{p}(c_i) = \frac{\sum_{k=1}^K p_{ik}}{K},
\end{equation}
where $\{p_{ik}\}_{k=1}^K$ is the set of retrieved prototypes for the given category $c_i$.

\tit{Superpixel-based Local Regions}
Once a visual representation of a class has been obtained through the aforementioned procedure, a straightforward solution to predict a segmentation mask for an image $I$ would be computing the semantic correspondences (\ie, cosine similarities) for each of its dense feature $v(I)$ against the representative prototypes of input categories $\overline{p}(I, c_i), \quad i = 1, \dots, S $, and interpolate the result to the original image size. However, such an approach would lead to noisy segmentation masks. 

In particular, it has been observed that DINOv2 shows good matching properties across objects from different images, but lacks in recognizing shapes and boundaries~\cite{zhang2023tale}. Hence, we propose to exploit a superpixel algorithm (\ie, the Felzenszwalb's algorithm~\cite{felzenszwalb2004efficient}) to partition the image by grouping pixels into class-agnostic non-overlapping regions according to their visual appearances and positions. 

Each superpixel can be interpreted as a binary mask $R \in \{0, 1\}^{H \times W}$ that is active on pixels belonging to it. Similar to the construction of visual prototypes, we interpolate each superpixel at the size of the dense features and perform a region pooling stage as defined in Eq.~\ref{eq:region_pooling} to produce superpixel embeddings $r_i \in \mathbb{R}^{D_v}, \quad i=1, \dots, |R|$. Then, for each superpixel embedding, we compute the cosine similarity against the representative prototypes of the categories. We associate each pixel with the unique region that includes it and we refer to this similarity in the unimodal space of the visual backbone as \textit{local similarity}.

\tit{Combining Local and Global Similarities}
While retrieved prototypes are linked with text, their feature vectors show good local matching properties but weaker global semantic capabilities. As correctly classifying pixels from a semantic point of view is crucial in segmentation, we propose to combine the local similarities obtained at the superpixel level with a global similarity measure which refers to the entire image.  We compute this in the multimodal space of a vision-language model (\ie, CLIP~\cite{radford2021learning}), which instead has good semantic classification capabilities. 

Specifically, we embed the input image using the image encoder of CLIP to produce an image embedding $i(I) \in \mathbb{R}^{D_t}$. Then, we compute cosine similarities between the image embedding and all the category embeddings $\hat{t}(c_i), \quad i=1, \dots, c_S$. Finally, we combine this global similarity with the single local similarities associated with class-agnostic regions. The final similarity between a local region and a semantic class is therefore computed as
\begin{equation}
    s(r_j, c_i) = \beta l(r_j, c_i) + (1-\beta)g(I, c_i),
\end{equation}
where $r_j$ indicates the local region, $c_i$ the semantic class, and $I$ the input image. Further, $l(r_j, c_i)$ is the local similarity between the region of interest and the class, and $g(I, c_i)$ is the global similarity extracted from CLIP space. To obtain the final segmentation mask, each region is then associated with the semantic class with the highest similarity.
\section{Experiments}
\label{sec:experiments}

\subsection{Experimental Setup}

\tit{Datasets}
We evaluate \ours on the validation splits of traditional semantic segmentation benchmarks, namely Pascal VOC 2012~\cite{pascal-voc-2012}, Pascal Context~\cite{mottaghi2014role}, COCO Stuff~\cite{caesar2018coco}, Cityscapes~\cite{cordts2016cityscapes}, and ADE20K~\cite{zhou2017scene,zhou2019semantic}. In particular, the validation sets of these datasets respectively contain 20, 59, 171, 150, and 19 semantic categories and 1,449, 5,104, 5,000, 2,000, and 500 images. In addition to these datasets for which we do not consider pixels not belonging to any category, we also validate our method when considering them as part of an additional ``unknown'' class (also referred to as ``background'' class in the literature). For these experiments, we again employ Pascal VOC 2012 and Pascal Context, and also include the COCO Objects dataset~\cite{caesar2018coco} which is a variant of COCO-Stuff with 80 foreground categories on the same validation split. To assess the segmentation performance, we employ the mean Intersection-over-Union (mIoU) on all the classes of each dataset.

\begin{table*}[t]
  \centering
  \setlength{\tabcolsep}{.4em}
  \resizebox{\textwidth}{!}{
  \begin{tabular}{lc c c cc cc c cc c ccccc}
    \toprule
     & & & & & & \multicolumn{2}{c}{\textbf{Parameters (M)}} & & \multicolumn{2}{c}{\textbf{Similarity}} & &  \multicolumn{5}{c}{\textbf{mIoU}} \\
    \cmidrule{7-8} \cmidrule{10-11} \cmidrule{13-17}
    \textbf{Model} & & \textbf{PAMR} & & \textbf{Dataset} & & Total & Trainable & & Textual & Visual & & VOC & Context & Stuff & Cityscapes & ADE\\
    \midrule
    ReCo~\cite{shin2022reco} & & \xmark & & ImageNet1k$^\bigstar$ & & 313.0 & 0.0 & & \xmark & \cmark & & 57.7 & 22.3 & 14.8 &           21.1 &  11.2\\
    GroupViT~\cite{xu2022groupvit} & & \xmark & & CC12M+RedCaps$^\blacklozenge$ & & 55.8 & 55.8 & & \cmark & \xmark & & 79.7 & 23.4 & 15.3 &  11.1 &   9.2 \\
    MaskCLIP~\cite{zhou2022extract} & & \xmark & & - & & 291.0 & 0.0 & & \cmark & \xmark & & 74.9 & 26.4 & 16.4 &             12.6 &   9.8 \\
    TCL~\cite{cha2023learning} & & \xmark & & CC3M+CC12M$^\blacklozenge$ & & 178.3 & 21.7 &  & \cmark & \xmark & & 77.5 & 30.3 & 19.6 &         23.1 &  14.9 \\
    OVDiff~\cite{karazija2023diffusion} & & \xmark & & - & & 1,226.4 & 0.0 & & \xmark & \cmark & & 81.7 & 33.7 & - &            -    &  14.9 \\
    \midrule                                                                                                              
    MaskCLIP~\cite{zhou2022extract} & & \cmark & & - & & 291.0 & 0.0 & & \cmark & \xmark & & 72.1 & 25.3 & 15.1 &             11.2 &   9.0 \\
    ReCo~\cite{shin2022reco} & & \cmark & & ImageNet1k$^\bigstar$ & & 313.0 & 0.0 & & \xmark & \cmark & & 62.4 & 24.7 & 16.3 &           22.8 &  12.4\\
    GroupViT~\cite{xu2022groupvit} & & \cmark & & CC12M+YFCC$^\blacklozenge$ & & 55.8 & 55.8 & & \cmark & \xmark & & 81.5 & 23.8 &            15.4  &  11.6 & 9.4  \\
    TCL~\cite{cha2023learning} & & \cmark & & CC3M+CC12M$^\blacklozenge$ & & 178.3 & 21.7  & & \cmark & \xmark & & 83.2 & 33.9 & 22.4 &        24.0 &  17.1 \\
    \midrule
     \rowcolor{Gray}
     \textbf{\ours (ViT-B)} & & \xmark & & COCO Captions$^\bigstar$ & & 236.1 & 0.0 & & \xmark & \cmark & & \textbf{85.6} (\textcolor{green}{\textbf{+2.4}}) & \textbf{43.1} (\textcolor{green}{\textbf{+9.2}}) & \textbf{27.8} (\textcolor{green}{\textbf{+5.4}}) & \textbf{36.7} (\textcolor{green}{\textbf{+12.7}}) & \textbf{22.4} (\textcolor{green}{\textbf{+5.3}})  \\
     \rowcolor{Gray} 
     \textbf{\ours (ViT-L)} & & \xmark & & COCO Captions$^\bigstar$ & & 732.0 & 0.0 & &\xmark & \cmark & & \textbf{87.9} (\textcolor{green}{\textbf{+4.7}}) & \textbf{43.5} (\textcolor{green}{\textbf{+9.6}}) & \textbf{28.8} (\textcolor{green}{\textbf{+6.4}}) & \textbf{36.7} (\textcolor{green}{\textbf{+12.7}}) & \textbf{23.2} (\textcolor{green}{\textbf{+6.1}}) \\
    \bottomrule
  \end{tabular}
  }
  \vspace{-0.15cm}
  \caption{Comparison with state-of-the-art unsupervised open-vocabulary semantic segmentation models on Pascal VOC~\cite{pascal-voc-2012}, Pascal Context~\cite{mottaghi2014role}, COCO Stuff~\cite{caesar2018coco}, Cityscapes~\cite{cordts2016cityscapes}, and ADE20K~\cite{zhou2017scene,zhou2019semantic}, without considering the unknown category. The markers $\blacklozenge$ and $\bigstar$ refer, respectively, to datasets used for training and support only.}
  \label{tab:model_comparison}
  \vspace{-0.4cm}
\end{table*}

\tit{Implementation Details}
Textual sentences used as input in our diffusion-augmented prototype generation pipeline are taken from the COCO Captions dataset~\cite{chen2015microsoft,lin2014microsoft}. We consider all five captions available for each image, thus obtaining a large set of captions describing natural images that can be used as input for a diffusion-based generative architecture. It is worth noting that we do not utilize the images associated with these captions. To generate the collection of visual prototypes, we employ Stable Diffusion v2.1~\cite{rombach2022high} with 50 diffusion steps and a threshold $\gamma$ equal to 0.45. The scalar weight $\alpha$ that combines the mean noun embeddings and caption embeddings to form keys is equal to 0.9.

We use DINOv2~\cite{oquab2023dinov2} pre-trained on the LVD-142M dataset as the self-supervised visual backbone, using both the ViT-B/14 and the ViT/L-14 versions, with an input image size of $518\times518$. This leads to dense features with size corresponding to $37\times37$. We also employ CLIP~\cite{radford2021learning} as the multimodal encoder using the original OpenAI weights, on top of the ViT-B/16 and ViT-L/14 architectures. We use the same CLIP model for both key embeddings and global similarity computation, so that \textit{(i)} we embed the arbitrary categories at inference time just one time and \textit{(ii)} we do not need to load two different text encoders into memory.

To extract superpixels, we use the Felzenszwalb's algorithm~\cite{felzenszwalb2004efficient}. We build and leverage an efficient exact retrieval index through the \texttt{faiss} library~\cite{johnson2019billion} based on cosine similarity. We consider the number of retrieved prototypes $K$ equal to 350 for all datasets and the ensembling weight $\beta$ between local and global similarities equal to 0.8 for all benchmarks except for Pascal VOC for which we use $\beta$ equal to 0.7.
More details are in the supplementary.

\tit{Evaluation Protocol}
To perform all experiments, we follow the unified evaluation protocol for unsupervised open-vocabulary semantic segmentation established by Cha~\etal~\cite{cha2023learning}. Specifically, we evaluate the model considering the class names from the default version of the \texttt{MMSegmentation} toolbox. We resize the images to have a shorter side equal to 448 and employ a sliding window approach with a stride of 224 pixels.

\subsection{Comparison with the State of the Art}
We first compare \ours with recent state-of-the-art approaches for unsupervised open-vocabulary semantic segmentation. Specifically, we include ReCo~\cite{shin2022reco} and OVDiff~\cite{karazija2023diffusion} that, similarly to our approach, exploit the arbitrary input categories to obtain a set of visual references. While ReCo curates an archive based on ImageNet1k~\cite{deng2009imagenet}, OVDiff generates a set of synthetic references at inference time by conditioning on a fixed prompt template, without necessitating external support data. Also, we compare with MaskCLIP~\cite{zhou2022extract}, which introduces some modifications to the CLIP architecture to exploit its multimodal embedding space, and GroupViT~\cite{xu2022groupvit} and TCL~\cite{cha2023learning} that rely on extensive contrastive training on large-scale datasets to learn a textual-visual alignment. When considering segmentation benchmarks with the background class, we also include ViewCo~\cite{ren2023viewco}, SegCLIP~\cite{luo2023segclip}, and OVSegmentor~\cite{xu2023learning} that, analogously to GroupViT and TCL, are based on natural language supervision via contrastive learning paradigms.

\begin{table}[t]
  \centering
  \setlength{\tabcolsep}{0.1em}
  \resizebox{\linewidth}{!}{
  \begin{tabular}{lcc cc ccc}
    \toprule
    & & & & & \multicolumn{3}{c}{\textbf{mIoU}} \\
    \cmidrule{6-8}
    \textbf{Model} & & \textbf{PAMR} & \textbf{Training Dataset} & & VOC & Context & Object \\
    \midrule
    GroupViT~\cite{xu2022groupvit} & & - & CC12M+RedCaps & & 50.4 & 18.7 & 27.5 \\
    MaskCLIP~\cite{zhou2022extract} & & - & - & & 38.8 & 23.6 & 20.6 \\
    ReCo~\cite{shin2022reco} & & - & - & & 25.1 & 19.9 & 15.7 \\
    ViewCo~\cite{ren2023viewco} & & - & CC12M+YFCC & & 52.4 & 23.0 & 23.5 \\
    SegCLIP~\cite{luo2023segclip} & & - & CC3M+COCO Captions & & 52.6 & 24.7 & 26.5 \\
    TCL~\cite{cha2023learning} & & - & CC3M+CC12M & & 51.2 & 24.3 & 30.4 \\
    OVSegmentor~\cite{xu2023learning} & & - & CC4M & & 53.8 & 20.4 & 25.1 \\
    \midrule
    GroupViT~\cite{xu2022groupvit} & & \cmark & CC12M+YFCC & & 51.1 & 19.0 & 27.9 \\
    MaskCLIP~\cite{zhou2022extract} & & \cmark & - & & 37.2 & 22.6 & 18.9 \\
    TCL~\cite{cha2023learning} & & \cmark & CC3M+CC12M & &
    55.0 & 30.4 & 31.6 \\
    \midrule
    \rowcolor{Gray}
    \textbf{\ours (ViT-L)} & & - & - & & \textbf{55.4} (\textcolor{green}{\textbf{+0.4}}) & \textbf{38.3} (\textcolor{green}{\textbf{+7.9}}) & \textbf{37.4} (\textcolor{green}{\textbf{+5.8}}) \\
    \bottomrule
  \end{tabular}
  }
  \vspace{-0.15cm}
  \caption{Comparison with state-of-the-art unsupervised open-vocabulary semantic segmentation models on the validation sets of Pascal VOC~\cite{pascal-voc-2012}, Pascal Context~\cite{mottaghi2014role}, and COCO Object~\cite{caesar2018coco}, when considering the additional unknown category. }
  \label{tab:model_comparison_bgr}
  \vspace{-0.5cm}
\end{table}

Table~\ref{tab:model_comparison} shows the results on the five benchmarks without the unknown category (\ie, Pascal VOC, Pascal Context, COCO Stuff, Cityscapes, and ADE20K). We report the performance of two variants of our approach: one based on DINOv2 ViT-B/14 and CLIP ViT-B/16 and the other based on DINOv2 ViT-L/14 and CLIP ViT-L/14, respectively denoted as \ours (ViT-B) and \ours (ViT-L). For this comparison, since the usage of superpixels to improve the adherence of predictions on the image can be interpreted as a mask refinement step, we also report the performance of considered competitors when using the Pixel-Adaptive Mask Refinement (PAMR) proposed in~\cite{araslanov2020single} to refine the final predictions. As it can be seen, both variants of our solution achieve the best results on all datasets, surpassing all the competitors by a consistent margin. Specifically, when comparing with methods without PAMR, \ours achieves an average improvement of 10.0 and 10.9 mIoU points with respect to TCL~\cite{cha2023learning}, respectively for the ViT-B and ViT-L variants. This performance improvement is confirmed also when comparing \ours with PAMR-based approaches, leading to an average increase of 7.0 and 7.9 mIoU points compared to the best-performing method. 

In Table~\ref{tab:model_comparison_bgr}, we instead report the results on the three segmentation datasets, namely Pascal VOC, Pascal Context, and COCO Object, used to validate the effectiveness of segmentation methods when also considering the additional ``unknown'' category. Following~\cite{xu2022groupvit}, we apply a threshold on the final similarities to detect pixels that do not belong to any of the provided input categories. In particular, we apply the threshold on the similarity values obtained after ensembling local and global similarities. For this experiment, we restrain the comparison to methods that do not employ specific techniques to take into account the background of the scene but instead perform a threshold as done in our case. Notably, \ours achieves the best results on all three benchmarks, surpassing both methods that do not employ any mask refinement stages and approaches that instead refine their predictions using PAMR~\cite{araslanov2020single}. In particular, \ours reaches 55.4, 38.3, and 37.4 mIoU points respectively on Pascal VOC, Pascal Context, and COCO Object, which correspond to an improvement of 0.4, 7.9, and 5.8 points with respect to the best method (\ie, TCL~\cite{cha2023learning} using PAMR as mask refinement technique).

These results highlight the effectiveness of our solution which, despite being completely training-free, achieves a new state of the art for unsupervised open-vocabulary semantic segmentation on all eight considered benchmarks. Some qualitative results are shown in Figure~\ref{fig:qualitatives}.

\subsection{Ablation Studies and Analyses}
We then evaluate the contribution of each component employed in our final solution and the effectiveness of different backbones to extract visual and textual features.

\begin{table}[t]
  \centering
  \setlength{\tabcolsep}{0.28em}
  \resizebox{\linewidth}{!}{
  \begin{tabular}{lc cc c ccc}
    \toprule
    & & & & & \multicolumn{3}{c}{\textbf{mIoU}} \\
    \cmidrule{6-8}
    \textbf{Backbone} & & \textbf{Global Similarity} & \textbf{Superpixels} & & VOC & Cityscapes & ADE \\
    \midrule
    CLIP (ViT-B/16) & & \xmark & \xmark & & 61.3 & 21.3 & 13.4 \\
    DINO (ViT-B/16) & & \xmark & \xmark & & 34.2 & 26.0 & 9.5 \\
    DINOv2 (ViT-B/14) & & \xmark & \xmark & & 75.6 & 34.4 & 20.7 \\
    \midrule
    DeiT-III (ViT-L/16) & & \xmark & \xmark & & 54.8 & 21.8 & 11.4 \\
    CLIP (ViT-L/14) & & \xmark & \xmark & & 45.9 & 20.0 & 11.4 \\
    DINOv2 (ViT-L/14) & & \xmark & \xmark & & 70.2 & 33.2 & 19.5 \\
    \midrule
    DINO (ViT-B/16) & & \cmark & \xmark & & 80.4 & 27.8 & 16.5 \\
    DINOv2 (ViT-B/14) & & \cmark & \xmark & & 86.2 & 35.0 & 21.9 \\
    DINOv2 (ViT-L/14) & & \cmark & \xmark & & 87.2 & 34.5 & 21.6 \\
    \midrule
    DINO (ViT-B/16) & & \cmark & \cmark & & 81.1 & 29.8 & 17.3 \\
    DINOv2 (ViT-B/14) & & \cmark & \cmark & & 87.0 & 36.6 & \textbf{23.2} \\
    \rowcolor{Gray} 
    DINOv2 (ViT-L/14) & & \cmark & \cmark & & \textbf{87.9} & \textbf{36.7} & \textbf{23.2} \\
    \bottomrule
  \end{tabular}
  }
  \vspace{-.15cm}
  \caption{Ablation study results using different visual backbones and validating the contribution of the key components of our solution. Results are reported on the validation sets of Pascal VOC~\cite{pascal-voc-2012}, Cityscapes~\cite{cordts2016cityscapes}, and ADE20K~\cite{zhou2017scene,zhou2019semantic}.}
  \label{tab:ablation_backbone}
  \vspace{-.4cm}
\end{table}

\tit{Effect of Changing the Visual Backbone}
We first consider the performance of our approach when using different visual backbones to compute local similarities. In particular, we evaluate DeiT-III~\cite{touvron2022deit} pre-trained for image classification on ImageNet1k and based on ViT-L/16, CLIP~\cite{radford2021learning} in both its ViT-B/16 and ViT-L/14 versions, DINO~\cite{caron2021emerging} based on the ViT-B/16 architecture, and our final choice DINOv2~\cite{oquab2023dinov2} using both the variant based on ViT-B/14 and the one based on ViT-L/14. Given that different input and patch sizes can lead to different output feature sizes, we resize all images to $518\times518$ when using visual backbones with a patch size of 14 and $592\times592$ when employing visual backbones with a path size of 16, thus always having features with a spatial size equal to $37\times37$. To validate only the role of different visual backbones, we apply them without global similarities and without superpixels to extract mask proposals. When considering the variant without superpixels, we directly compute the local similarities on the dense features and we interpolate them to the original image size.

\begin{table}[t]
  \centering
  \setlength{\tabcolsep}{0.3em}
  \resizebox{\linewidth}{!}{
  \begin{tabular}{lcc c ccc}
    \toprule
    & & & & \multicolumn{3}{c}{\textbf{mIoU}} \\
    \cmidrule{5-7}
    \textbf{Local Backbone} & & \textbf{Textual/Global Backbone} & & VOC & Cityscapes & ADE \\
    \midrule
    DINO (ViT-B/16) & & CLIP (ViT-B/16) & & 80.8 & 30.6 & 17.0 \\
    DINOv2 (ViT-B/14) & & CLIP (ViT-B/16) & & 85.6 & \textbf{36.7} & 22.4 \\
    DINOv2 (ViT-L/14) & & CLIP (ViT-B/16) & & 86.9 & 36.3 & 22.3 \\
    \midrule
    \rowcolor{Gray} 
    DINOv2 (ViT-L/14) & & CLIP (ViT-L/14) & & \textbf{87.9} & \textbf{36.7} & \textbf{23.2} \\
    \bottomrule
  \end{tabular}
  }
  \vspace{-.15cm}
  \caption{Performance analysis when employing visual and textual backbones of different sizes.}
  \label{tab:ablation_size}
  \vspace{-.4cm}
\end{table}

\begin{figure*}[t]
\setlength{\tabcolsep}{.2em}
\includegraphics[width=0.99\linewidth]{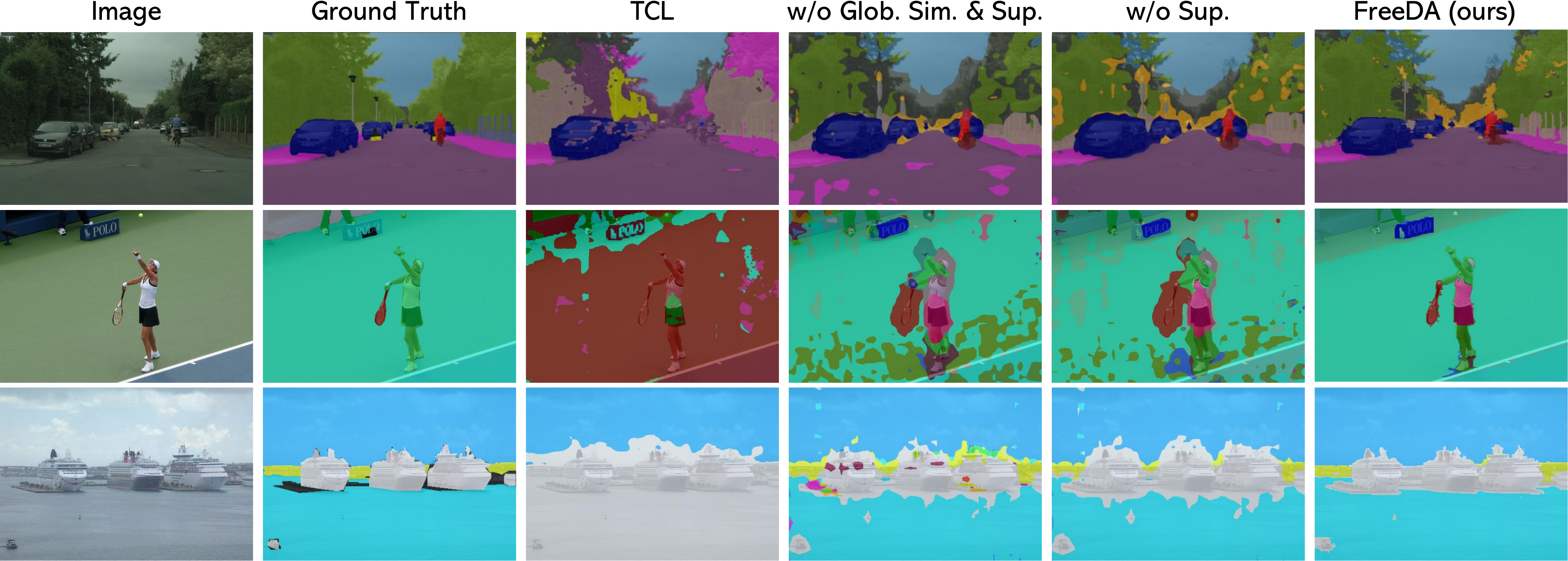}
\vspace{-.15cm}
\caption{Qualitative results of \ours in comparison with TCL~\cite{cha2023learning}, with and without global similarities and superpixels.}
\label{fig:qualitatives}   
\vspace{-.45cm}
\end{figure*}

Results are reported in the upper part of Table~\ref{tab:ablation_backbone}, using the CLIP ViT-L/14 model to extract textual features. As it can be noticed, DINOv2 exhibits the best performance among both architectures based on ViT-B and ViT-L, confirming the power of self-supervised features in this setting.

\tit{Adding Global Similarities and Superpixels}
To evaluate the contribution of global features and superpixel-based mask proposals, we report in the lower part of Table~\ref{tab:ablation_backbone} the performance of \ours first adding only global similarities and then also including superpixels to extract mask proposals. Both strategies give a consistent contribution to the final performance, also when considering different visual backbones to compute local similarities. For example, when using DINOv2, global features bring an improvement of 0.9 mIoU points on the ADE20K dataset, while superpixels further enhance the final performance by an additional 1.6 mIoU points. Additionally, it is worth noting that the contribution of global similarities is more significant in Pascal VOC where images are characterized by the presence of a single or few objects occupying large areas of the scene, thus favoring global features instead of local ones.

\tit{Impact of Backbone Size} In Table~\ref{tab:ablation_size} we investigate how much using a ViT-Large architecture to extract both visual and textual features increases the performance compared to a ViT-Base model. As also demonstrated by the complete results of the two variants of \ours reported in Table~\ref{tab:model_comparison}, this corresponds to around 2.3 mIoU points on Pascal VOC when employing DINOv2 to extract local features, while obtaining similar performance on Cityscapes and ADE20K.

\tit{Superpixel Algorithms and Prototype Aggregation Strategies} In Table~\ref{tab:ablation}, we instead validate the choice of employing Felzenszwalb's algorithm~\cite{felzenszwalb2004efficient} to extract superpixels by comparing it with three widely adopted superpixel proposal algorithms, namely Watershed~\cite{hu2015watershed}, SLIC~\cite{achanta2012slic}, and SEEDS~\cite{van2012seeds}. While different versions of superpixel algorithms lead to similar performance, the usage of Felzenszwalb's algorithm helps to further improve the results on all three datasets considered. In addition to comparing different superpixel extraction strategies, we also include the results obtained using PAMR~\cite{araslanov2020single} as a mask refinement method. For this experiment, we first compute local similarities for dense features and ensemble them with the global similarity, then we apply PAMR to refine the resulting segmentation masks. Notably, employing superpixels to extract mask proposals leads to improved final results.

To validate the aggregation strategy used in \ours, in which we aggregate retrieved prototypes by computing their average embedding (\ie, ``mean embedding'' in Table~\ref{tab:ablation}), we compare it with two different approaches based on first computing local similarities for all retrieved prototypes and then aggregating them by considering the mean or the maximum (\ie, ``mean similarity'' and ``max similarity''). Computing the average embedding of all retrieved prototypes brings the best results across all datasets.

\begin{table}[t]
  \centering
  \setlength{\tabcolsep}{0.32em}
  \resizebox{\linewidth}{!}{
  \begin{tabular}{lcc c ccc}
    \toprule
    & & & & \multicolumn{3}{c}{\textbf{mIoU}} \\
    \cmidrule{5-7}
    \textbf{Model} & & \textbf{Superpixels} & & VOC & Cityscapes & ADE \\
    \midrule
    w/ mean embedding (PAMR) & & - & & 87.0 & 34.4 & 23.0 \\
    \midrule
    w/ mean embedding & & Watershed & & 87.0 & 32.7 & 21.8 \\
    w/ mean embedding & & SLIC & & 87.3 & 33.5 & 21.8 \\
    w/ mean embedding & & SEEDS & & 87.5 & 32.3 & 22.4 \\
    \midrule
    w/ mean similarity & & Felzenszwalb & & 79.5 & 29.3 & 18.8 \\
    w/ max similarity & & Felzenszwalb & & 82.0 & 26.2 & 17.6 \\
    \rowcolor{Gray} 
    \textbf{\ours} (w/ mean embedding) & & Felzenszwalb & & \textbf{87.9} & \textbf{36.7} & \textbf{23.2} \\
    \bottomrule
  \end{tabular}
  }
  \vspace{-.15cm}
  \caption{Performance analysis using different algorithms to compute superpixels and different prototypes aggregation strategies.}
  \label{tab:ablation}
  \vspace{-.45cm}
\end{table}

\begin{figure}[t]
\setlength{\tabcolsep}{.3em}
\resizebox{\linewidth}{!}{
\begin{tabular}{cc}
\includegraphics[height=0.49\linewidth]{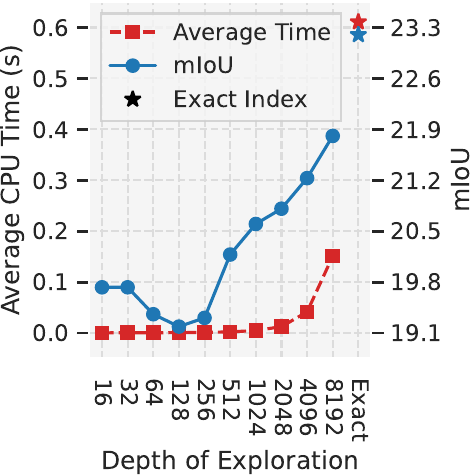} & 
\includegraphics[height=0.49\linewidth]{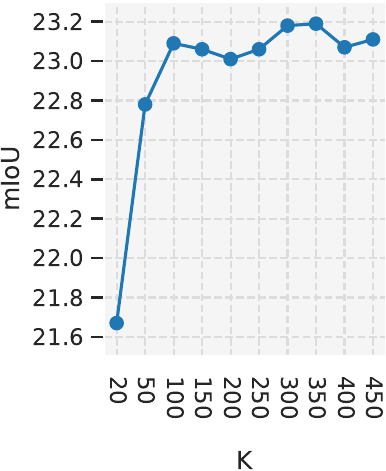} 
\end{tabular}
}
\vspace{-.25cm}
\caption{Retrieval results when using an approximate index (left) and varying the number of retrieved key-prototype pairs (right).}
\label{fig:ablation}   
\vspace{-.45cm}
\end{figure}

\tit{Retrieval Performance Analysis}
Finally, we analyze the performance when varying the retrieval parameters. Since our method leverages an exact retrieval index, we first validate how much using an approximate search impacts the performance. Specifically, the left plot of Figure~\ref{fig:ablation} shows the trade-off between speed and performance when using a graph-based HNSW (Hierarchical Navigable Small World) index~\cite{malkov2018efficient}. We report the CPU times to search the most similar $K=350$ key embeddings when changing the depth of exploration in the index, and their corresponding mIoU scores. This parameter controls the size of the dynamic list of candidate nearest neighbors that are explored during the search process. On the right plot of Figure~\ref{fig:ablation}, we instead show the performance variation when changing the number $K$ of searched keys. Results are reported on the ADE20K dataset. As it can be seen, using an approximate index only partially deteriorates the performance while consistently reducing time computation. On the same line, increasing the number of retrieved key embeddings does not improve the final performance, while retrieving a reduced number of items partially leads to lower results. 

\section{Conclusion}
\label{sec:conclusion}
We presented \ours, a training-free approach for unsupervised open-vocabulary segmentation. Our approach leverages visual prototypes and textual keys extracted offline with diffusion-augmented generation and exploits local-global similarities at inference time. Experimentally, we achieve state-of-the-art results on five different datasets.

%



\section*{Acknowledgment\vspace{-0.1cm}}
This work has been conducted under a research grant co-funded by Leonardo S.p.A., and supported by the PNRRM4C2 project ``FAIR - Future Artificial Intelligence Research'' and by the EU Horizon project ``ELIAS - European Lighthouse of AI for Sustainability'' (No. 101120237), both funded by the European Commission.

{
\small
\bibliographystyle{ieeenat_fullname}
\bibliography{bibliography}
}

\clearpage
\appendix

In this supplementary material, we delve into additional implementation details pertaining to our prototype generation process, offering information to facilitate reproducibility. A comprehensive list of the used textual prompts is presented to clarify the experimental setup. We systematically explore the impact of varying superpixel hyperparameters on the overall performance of our proposed model.
We examine the combined influence of entire caption contexts and word embeddings during prototype generation. Our findings highlight the effectiveness of this approach, particularly for categories consisting of multiple words. 
We also investigate the impact of employing the unimodal backbone for both local and global matching. Our results demonstrate the advantage of leveraging a multimodal feature extractor like CLIP for global matching.
To enhance interpretability, we include visual examples showcasing captions, generated images, and their corresponding attributions and binary masks.
Additionally, we include qualitative results across all the considered benchmark datasets. We conduct a thorough examination of both successful cases and instances of failure, supplementing our analysis with ``into the wild'' examples—segmentation results obtained by prompting our model with diverse free-form textual inputs.

\section{Additional Implementations Details}
\tinytit{Textual Templates}
To encode through the CLIP text encoder both the nouns extracted during prototype generation and the input categories utilized at inference time, we employ the following set of templates $\mathcal{T}$, introduced in~\cite{radford2021learning}:
\begin{itemize}
    \item[] \texttt{itap of a \{\}.}
    \item[] \texttt{a bad photo of the \{\}.}
    \item[] \texttt{a origami \{\}.}
    \item[] \texttt{a photo of the large \{\}.}
    \item[] \texttt{a \{\} in a video game.}
    \item[] \texttt{art of the \{\}.}
    \item[] \texttt{a photo of the small \{\}.}
\end{itemize}
As discussed in~\cite{radford2021learning}, these templates provide a powerful means of contextualizing textual input, making them particularly well-suited for our application in the context of prototype generation and inference.

\tit{Prototypes generation}
The foundation of our prototype generation lies in the utilization of a dataset of images paired with captions.
To ensure the reproducibility of our results, we detail the negative prompts employed during the generation of images with Stable Diffusion in Table~\ref{tab:negative_prompts}. These negative prompts play a crucial role in guiding the generation process, aiming to produce prototypes that are realistic and high-quality. 
The prototypes generation is performed offline and requires around 5.2 sec for each COCO caption. During inference, computing a category embedding and performing prototypes retrieval takes around 10.8 ms and 12.9 ms for the Base and Large versions of \ours.

\begin{table}[t]
\centering
\resizebox{\linewidth}{!}{
    \begin{tabular}{l l l}
    \toprule
\rowcolor{lightgray} \emph{3d}
& \emph{abstract} & \emph{art} \\
\emph{asymmetric}
& \emph{bad anatomy} & \emph{bad art} \\
\rowcolor{lightgray} \emph{bad proportions}
& \emph{blurry} & \emph{canvas frame} \\
\emph{cartoon}
& \emph{cartoonish} & \emph{cgi} \\
\rowcolor{lightgray} \emph{cloned face}
& \emph{colorless} & \emph{computer graphic} \\
\emph{cropped}
& \emph{cut off} & \emph{deformed} \\
\rowcolor{lightgray} \emph{dehydrated}
& \emph{digital} & \emph{digital art} \\
\emph{disfigured}
& \emph{doll} & \emph{duplicate} \\
\rowcolor{lightgray} \emph{error}
& \emph{extra arms} & \emph{extra fingers} \\
\emph{extra legs}
& \emph{extra limbs} & \emph{fused fingers} \\
\rowcolor{lightgray} \emph{fuzzy}
& \emph{grainy} & \emph{graphic} \\
\emph{gross proportions}
& \emph{inaccurate} & \emph{jpeg artifacts} \\
\rowcolor{lightgray} \emph{long neck}
& \emph{low quality} & \emph{low-resolution} \\
\emph{lowres}
& \emph{malformed limbs} & \emph{misshaped} \\
\rowcolor{lightgray} \emph{missing arms}
& \emph{missing legs} & \emph{morbid} \\
\emph{mutant}
& \emph{mutated} & \emph{mutated hands} \\
\rowcolor{lightgray} \emph{mutation}
& \emph{mutilated} & \emph{octane} \\
\emph{out of focus}
& \emph{out of frame} & \emph{oversaturated} \\
\rowcolor{lightgray} \emph{photoshop}
& \emph{poorly drawn face} & \emph{poorly drawn hands} \\
\emph{render}
& \emph{retro} & \emph{signature} \\
\rowcolor{lightgray} \emph{text}
& \emph{too many fingers} & \emph{ugly} \\
\emph{unreal}
& \emph{unreal engine} & \emph{unrealistic} \\
\rowcolor{lightgray} \emph{username}
& \emph{video game} & \emph{watermark} \\
\emph{weird colors}
& \emph{worst quality} \\
    \bottomrule
    \end{tabular}
    }
    \vspace{-0.1cm}
    \caption{Negative prompts employed in Stable Diffusion during prototypes generation.}
    \label{tab:negative_prompts}
\vspace{-0.15cm}
\end{table}

\section{Additional Experiments and Analyses}
\label{sec:additional_experiments}

\tit{Effect of Superpixel Parameters}
Felzenszwalb~\etal~\cite{felzenszwalb2004efficient} introduced an efficient superpixel algorithm that employs a graph-based approach. The algorithm initiates by constructing a graph representation of the image, where each pixel serves as a node, and edges connect neighboring pixels. Edge weights are determined based on the RGB color space differences between adjacent pixels. Consequently, connected components, initially established as individual components for each pixel, are progressively merged. The growth of each component is regulated by the scale of observation parameter $k$. The algorithm also incorporates two additional parameters: the diameter of the Gaussian filter used for pre-processing to enhance image smoothness and counter artifacts ($\sigma$), and the enforced minimum size of superpixels, $\mu$. We employ the implementation of the \texttt{skimage}\footnote{\url{https://scikit-image.org/}} library.

In Table~\ref{tab:superpixel_params}, we report the parameter values employed on the examined datasets. Figure~\ref{fig:superpixel_graph_supp} further shows the performance variations obtained when altering these parameters on the ADE20K dataset~\cite{zhou2017scene,zhou2019semantic}. Notably, minor variations in these parameters have negligible effects on final performance. However, imposing large superpixels through minimum size or scale of observation can significantly degrade the results.

\begin{table}[t]
  \centering
  \setlength{\tabcolsep}{0.5em}
  \resizebox{0.55\linewidth}{!}{
  \begin{tabular}{l ccc}
    \toprule
    \textbf{Dataset} & \textbf{$\mu$} & \textbf{$\sigma$} & \textbf{$k$} \\
    \midrule
    Pascal VOC & 100 & 0.7 & 20 \\
    Pascal Context & 100 & 1.0 & 20 \\
    COCO Stuff & 100 & 1.0 & 100 \\
    Cityscapes & 50 & 0.5 & 20 \\
    ADE20K & 100 & 1.0 & 20 \\
    \bottomrule
  \end{tabular}
  }
  \vspace{-.1cm}
  \caption{Parameters employed for Felzenszwalb's algorithm on each dataset.}
  \label{tab:superpixel_params}
  \vspace{-.1cm}
\end{table}

\begin{figure}[t]
\setlength{\tabcolsep}{.05em}
\resizebox{\linewidth}{!}{
\centering
\begin{tabular}{ccc}
\includegraphics[height=0.4\linewidth]{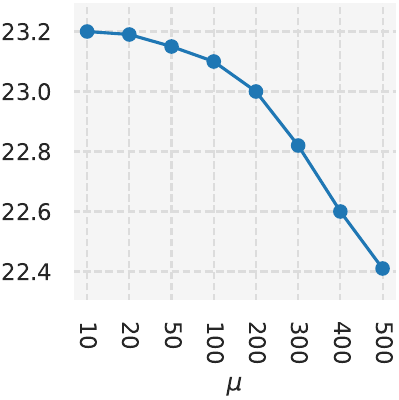} & 
\includegraphics[height=0.4\linewidth]{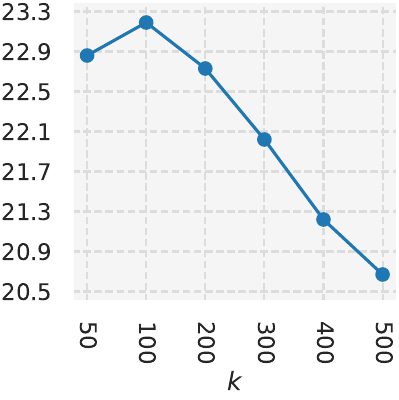} &
\includegraphics[height=0.4\linewidth]{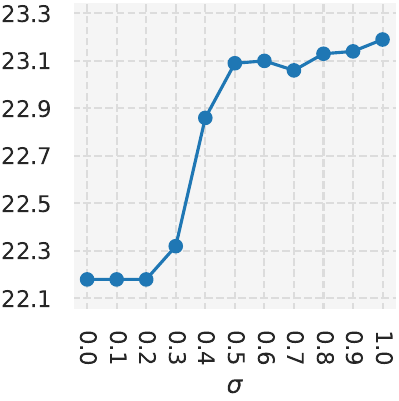}
\end{tabular}
}
\vspace{-.15cm}
\caption{Effect of the variation of superpixel hyperparameters on ADE20K, measured in terms of mIoU.}
\label{fig:superpixel_graph_supp}   
\vspace{-.3cm}
\end{figure}

\tit{Impact of caption context}
In Section~\ref{sec:prototype-generation} of the main paper, we outline our methodology for extracting textual key embeddings. Specifically, we employ a linear combination of the word embedding $\hat{t}$ and the caption embedding $\hat{c}$, controlled by a parameter $\alpha$. In our main results, we set $\alpha$ to 0.9 to effectively incorporate the textual context into the key embedding.

In Table~\ref{tab:caption_context_supp}, we conduct an ablation study on this choice. The case without caption context corresponds to setting $\alpha$ to 1. It is noteworthy that the inclusion of textual context proves to be particularly beneficial for input categories that consist of more than one word, such as \texttt{chest of drawers}. This scenario is prevalent in in-the-wild situations, thus emphasizing the practical utility of our approach in diverse and real-world settings.

\tit{Impact of unimodal global matching}
In Table~\ref{tab:ablation_single_backbone}, we investigate the impact of employing DINOv2 for local and global matching. 
Since DINOv2 embeddings are not aligned with text, we compute global matching by using the similarity between the \texttt{\small CLS} token of DINOv2 and the representative visual prototypes of the categories.
As can be observed, the usage of a text-aligned CLIP backbone improves performance \wrt the unimodal DINOv2 global features.

\section{Explainability}
A notable advantage of our prototype-based approach lies in its inherent explainability, as the set of referring images used to generate prototypes can be visualized a posteriori. In our approach, in particular, we can visualize the generated images associated with the retrieved prototypes for a given input category, along with the corresponding attribution maps and binary masks.

\begin{figure*}[t]
\centering
\includegraphics[width=1.0\linewidth]{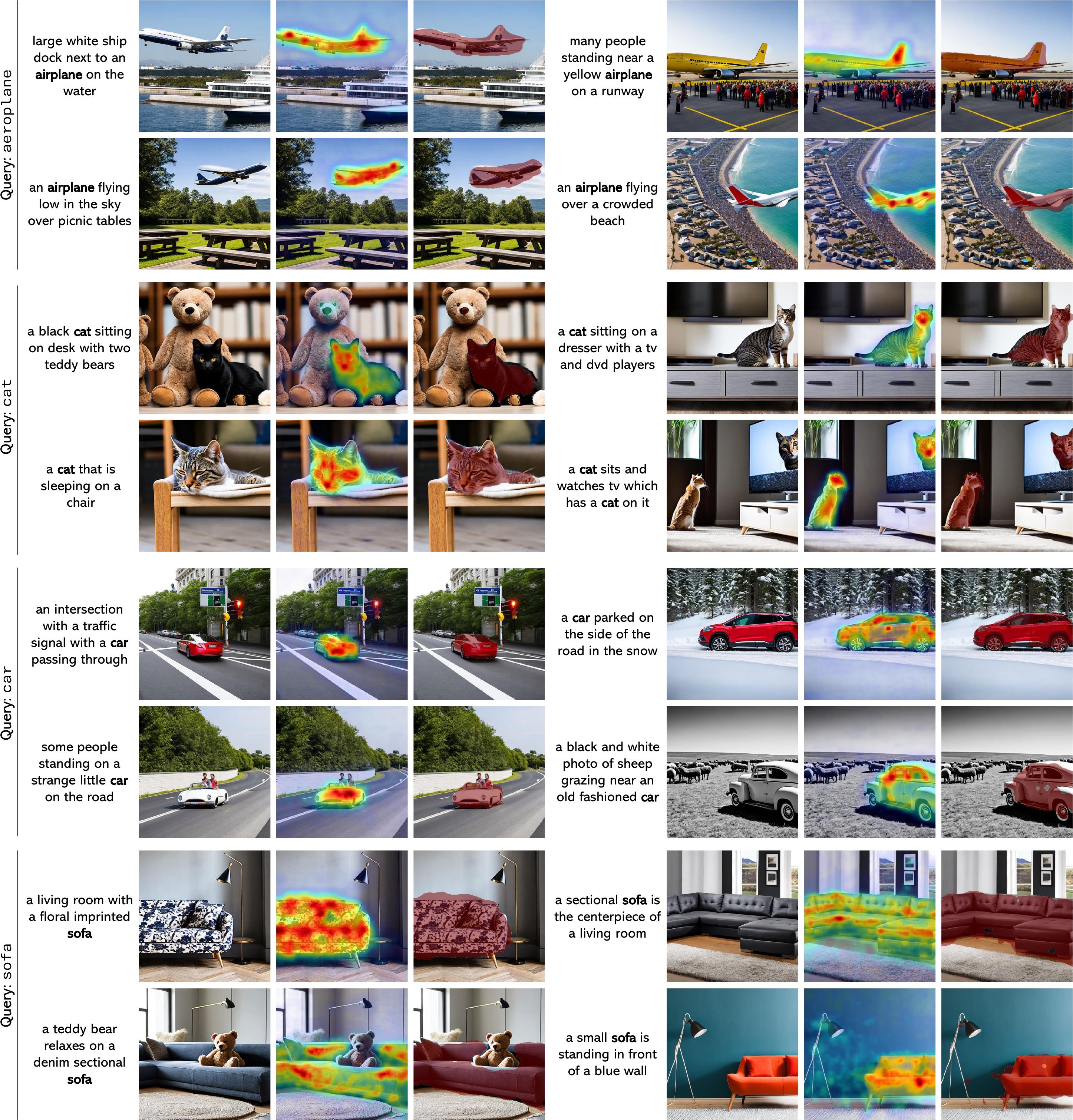}
\caption{Examples of retrieved prototypes for a specified textual category. From left to right, we show the original COCO caption, the corresponding generated image, the attribution map, and the binarized mask (area highlighted in red).}
\label{fig:explainability}
\end{figure*}

Figure~\ref{fig:explainability} illustrates the explainability capabilities of our solution, showcasing examples of retrieved prototypes for a specified category, highlighted within the captions in which the corresponding noun was mentioned. We further include the corresponding generated images, attribution maps, and binarized masks, providing a comprehensive view of the explainability achieved by our approach.

\begin{table}[t]
  \centering
  \setlength{\tabcolsep}{0.5em}
  \resizebox{0.85\linewidth}{!}{
  \begin{tabular}{lcc ccc}
    \toprule
    & & & \multicolumn{3}{c}{\textbf{mIoU}} \\
    \cmidrule{4-6}
    & \textbf{Caption Context} & & Context & Stuff & ADE \\
    \midrule
    & \xmark & & 43.1 & 27.4 & 22.2 \\
    \rowcolor{Gray} 
    \textbf{\ours} & \cmark & & \textbf{43.5} & \textbf{28.8} & \textbf{23.2} \\
    \bottomrule
  \end{tabular}
  }
  \vspace{-.1cm}
  \caption{Effect of full caption embeddings on the performance of key embeddings.}
  \label{tab:caption_context_supp}
\end{table}

\begin{table}[t]
 \centering
        \resizebox{.99\linewidth}{!}{
            \begin{tabular}{lcc c ccc}
                \toprule
                \textbf{Local Backbone} & & \textbf{Global Backbone} & & VOC & Cityscapes & ADE \\
                \midrule
                DINOv2 (ViT-B/14) & & DINOv2 (ViT-B/14) & & 78.4 & 30.7 & 17.8 \\
                DINOv2 (ViT-L/14) & & DINOv2 (ViT-L/14) & & 74.4 & 33.5 & 20.3 \\
                \midrule
                \rowcolor{Gray} 
                DINOv2 (ViT-B/14) & & CLIP (ViT-B/16) & & 85.6 & \textbf{36.7} & 22.4 \\
                \rowcolor{Gray} 
                DINOv2 (ViT-L/14) & & CLIP (ViT-L/14) & & \textbf{87.9} & \textbf{36.7} & \textbf{23.2} \\
                \bottomrule
            \end{tabular}
        }
        \captionof{table}{mIoU results with DINOv2 for local/global matching.}
        \label{tab:ablation_single_backbone}
        \vspace{-0.3cm}
\end{table}

\begin{figure*}[t]
\centering
\setlength{\tabcolsep}{.2em}
\includegraphics[width=0.99\linewidth]{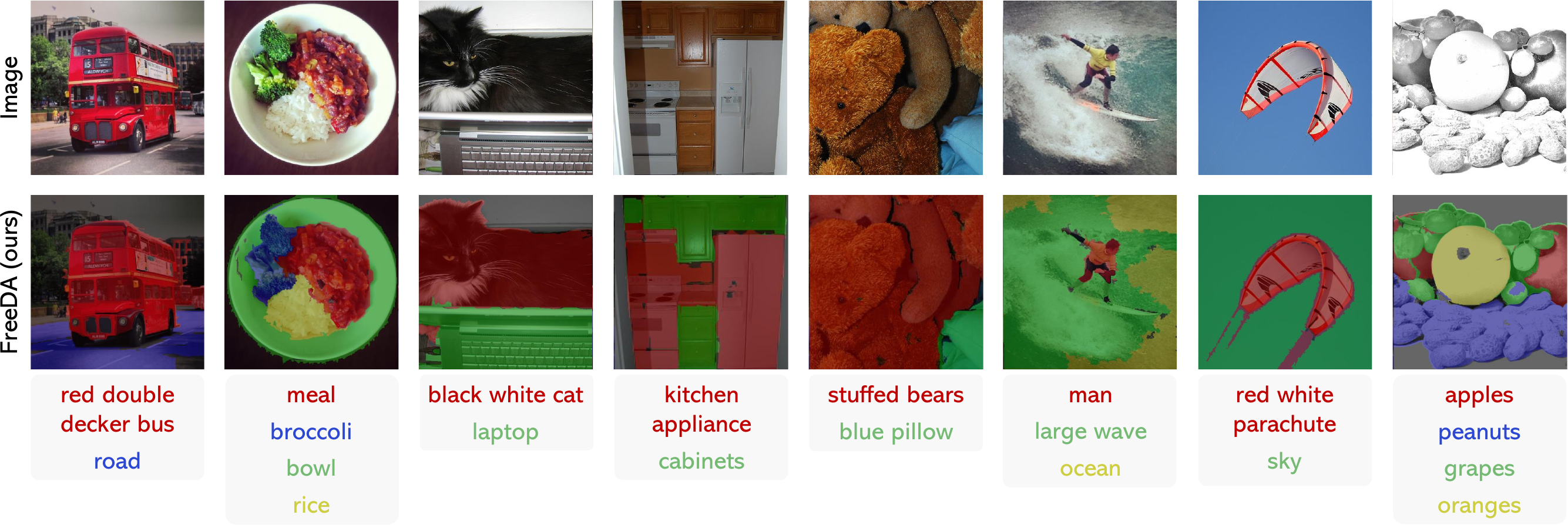}
\vspace{-.15cm}
\caption{In-the-wild segmentation results obtained by prompting our model with diverse free-form textual inputs.}
\label{fig:into_the_wild_supp}   
\vspace{-.1cm}
\end{figure*}

\begin{figure}[t]
\centering
\setlength{\tabcolsep}{.2em}
\includegraphics[width=0.99\linewidth]{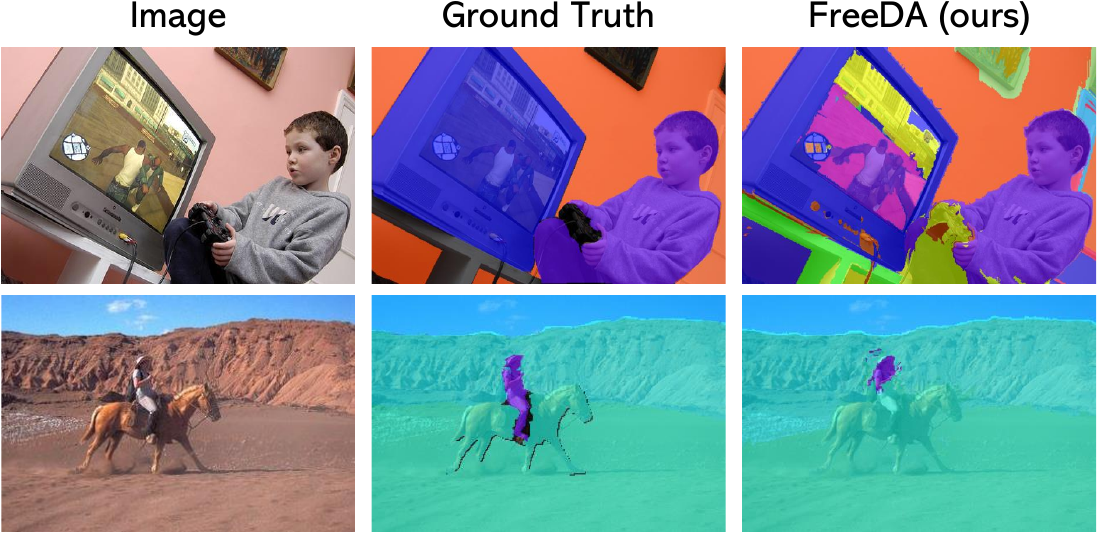}
\vspace{-.1cm}
\caption{Sample failure cases.}
\label{fig:failures_supp}   
\vspace{-.1cm}
\end{figure}

\section{Additional Qualitative Results}
\begin{figure*}[t]
\centering
\includegraphics[width=1.0\linewidth]{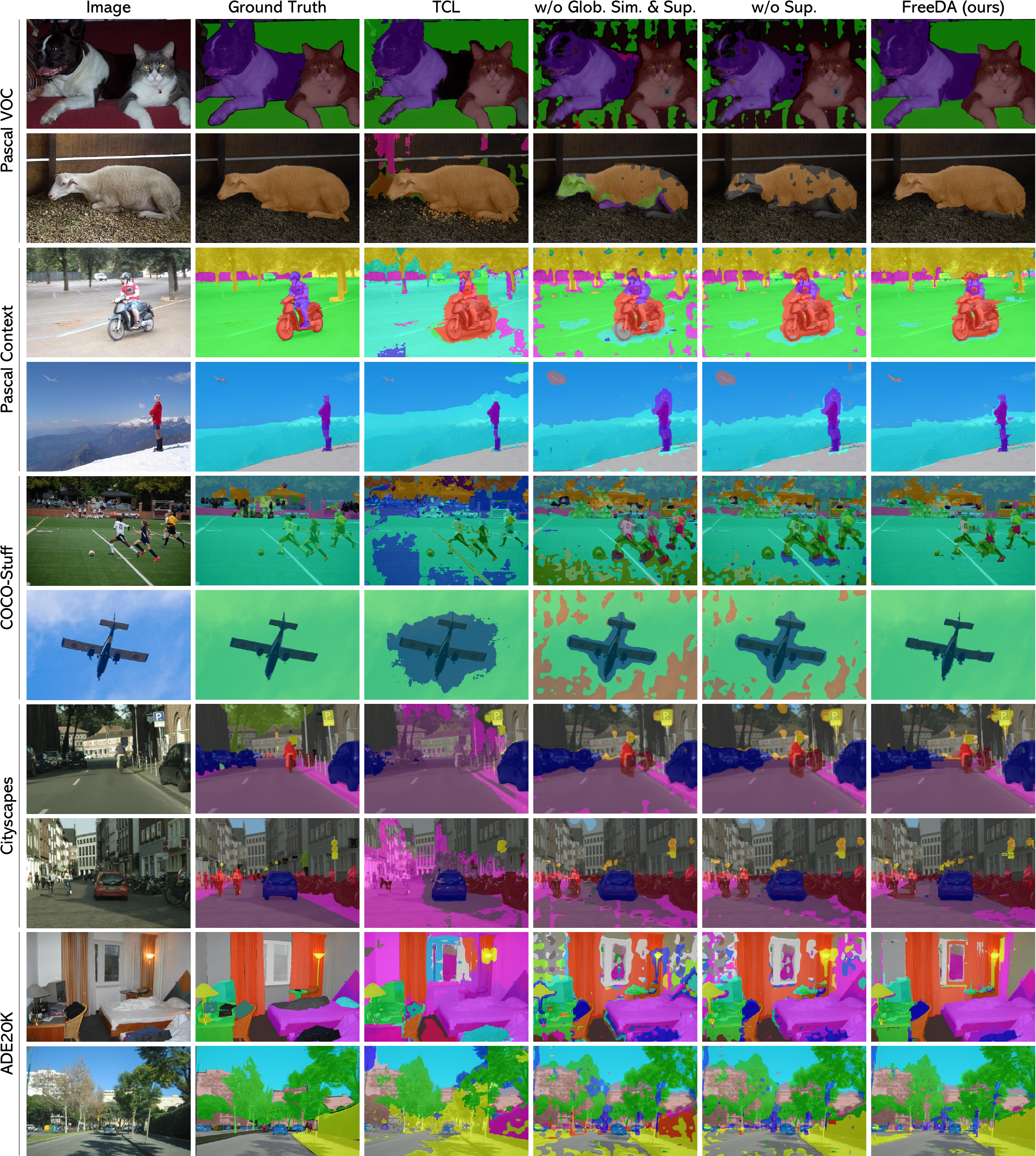}
\caption{Additional qualitative results of \ours in comparison with TCL~\cite{cha2023learning}, with and without global similarities and superpixels.}
\label{fig:more_qualitatives}   
\end{figure*}

\tit{Results on benchmark datasets}
Figure~\ref{fig:more_qualitatives} shwocases additional qualitative results on Pascal VOC~\cite{pascal-voc-2012}, Pascal Context~\cite{mottaghi2014role}, COCO Stuff~\cite{caesar2018coco}, Cityscapes~\cite{cordts2016cityscapes}, and ADE20K~\cite{zhou2017scene,zhou2019semantic}. These qualitative samples offer a comprehensive view of the performance of our approach, and highlight the versatility and effectiveness of our method across a range of scenes and categories, reinforcing its applicability in various real-world scenarios.

\tit{In-the-wild results}
Additionally, in Figure~\ref{fig:into_the_wild_supp} we report a collection of in-the-wild examples obtained by prompting our model with diverse free-form textual inputs. Specifically, we extract noun chunks from sample captions of the COCO Captions validation set using the \texttt{spaCy}\footnote{\url{https://spacy.io/}} NLP library. After removing stop-words, the noun chunks are utilized as input categories for segmenting the corresponding images. These results extend our analysis beyond curated datasets and demonstrate the adaptability and robustness of our approach in handling real-world scenarios with varied and unstructured textual descriptions.

\tit{Failure cases}
Finally, in Figure~\ref{fig:failures_supp} we report sample scenarios in which our model encounters challenges and exhibits failure cases. The first row illustrates an image of a TV displaying a video game. Owing to the strong semantic correspondence properties at the token-level of DINOv2, our model tends to segment individual elements shown on the TV screen, thereby impacting the overall segmentation performance for the TV class.
The second row of the figure instead presents another failure case featuring an image of a person atop a horse. However, the segmentation is incomplete and only partially captures the person. This limitation can be attributed to the prototypes corresponding to horses ridden by persons, whose noisy binarized masks include their legs.
Overall, these failure cases shed light on areas where our model may struggle, emphasizing the need for further refinement and consideration of complex visual contexts.

\end{document}